\documentclass[10pt,conference]{IEEEtran}
\IEEEoverridecommandlockouts

\usepackage[table]{xcolor}
\usepackage{xcolor}
\usepackage{cite}
\usepackage{makecell}
\usepackage{tabularx} 
\usepackage{enumitem}
\usepackage{array}
\usepackage{colortbl}
\usepackage{adjustbox}
\usepackage[linesnumbered,ruled,vlined]{algorithm2e}
\usepackage[utf8]{inputenc}
\usepackage{booktabs} 
\usepackage{amsmath,amssymb,amsfonts}
\usepackage{graphicx}
\usepackage{textcomp}
\usepackage{url}
\usepackage{titlesec}

\def\BibTeX{{\rm B\kern-.05em{\sc i\kern-.025em b}\kern-.08em
    T\kern-.1667em\lower.7ex\hbox{E}\kern-.125emX}}

\titlespacing\subsection{0pt}{3pt plus 1pt minus 1pt}{3pt plus 1pt minus 1pt}

\begin{document}

\title{Enhancing Forecasting with a 2D Time Series Approach for Cohort-Based Data}

\author{\IEEEauthorblockN{Yonathan Guttel}
\IEEEauthorblockA{\textit{Lightricks}\\
Jerusalem, Israel \\
Email: yguttel@lightricks.com}
\and
\IEEEauthorblockN{Orit Moradov}
\IEEEauthorblockA{\textit{Lightricks}\\
Jerusalem, Israel \\
Email: orit@lightricks.com}
\and
\IEEEauthorblockN{Nachi Lieder}
\IEEEauthorblockA{\textit{Lightricks}\\
Jerusalem, Israel \\
Email: nachi@lightricks.com}
\and
\IEEEauthorblockN{Asnat Greenstein-Messica}
\IEEEauthorblockA{\textit{Lightricks}\\
Jerusalem, Israel \\
Email: asi@lightricks.com}
}

\maketitle
\begin{abstract}
This paper introduces a novel two-dimensional (2D) time series forecasting model that integrates cohort behavior over time, addressing challenges in small data environments. We demonstrate its efficacy using multiple real-world datasets, showcasing superior performance in accuracy and adaptability compared to reference models. The approach offers valuable insights for strategic decision-making across industries facing financial and marketing forecasting challenges.
\end{abstract}
\vspace{1pt}

\begin{IEEEkeywords}
Time Series Forecasting, ARIMA, Cohort Analysis, User Subscription Behavior Analysis, Small Data Environments, Multidimensional Forecasting, Time Series Data Representation
\end{IEEEkeywords}

\section{Introduction}
In the dynamic landscape of time series forecasting, challenges persist in accurately predicting future events, particularly in environments characterized by limited data. This limitation is widespread in various business contexts, where the ability to forecast multiple steps ahead can significantly impact decision-making and strategic planning. Traditional time series models, while robust in data-rich scenarios \cite{1}, often struggle with the nuances of smaller datasets, where irregularities and non-linear trends are more pronounced.

In addition to these traditional concerns, there arises a more nuanced issue. A subclass of time series includes user based activity which can be represented both in a chronological way and a cohort-like manner. An example of this would be a user subscription-based time series. Users may subscribe to a service at any given time with a weekly/monthly/yearly subscription. This creates a more complex time series as we may experience inner-relations between two aspects, the cohort (e.g., group of users attributed to their installation time) and actual time the user subscribed or renewed. Building a time series model based on one of the aspects without addressing the other may result in inadequate performance.

Previous studies focused on either time series or cohort-level analysis separately \cite{2,3,4}. We aim to combine both approaches for tighter predictions. Our goal is to forecast cohort-level user behavior while incorporating chronological time elements. We explore whether we can effectively predict future time series observations using cohort-level dimensions in a small data environment.

This paper introduces a novel method to enhance predictive accuracy in small data environments by transforming the representation of the data. Our approach decomposes the multi-step forecasting problem into a two-dimensional representation. The dimensions will represent on one hand the cohort resolution, and on the other the horizon looking forward (the cohort behavior) - number of prediction steps since an event occurs (e.g., months since installation).
We will explain how these two aspects are not only related but also intertwine and assist each other in the prediction, as well as tighten the prediction error going forward.
We will also discuss how re-shaping the data can create an environment which may obey the model assumptions for time series models. Additionally, we will describe how predictions are implemented by filling in a two-dimensional matrix to construct a series of forecasts.

The source code is publicly available at \url{https://github.com/Lightricks/cohort-based-2D-time-series}.

\section{Related Work} In small data environments, understanding the performance of forecasting models under such constraints is crucial. In cohort-level forecasting, challenges arise from small sample sizes and lack of repetitive seasonal cycles, often necessitating simpler regression-based approaches that compromise accuracy. Robust models are needed to adapt to limited data while capturing time series dynamics. Autoregressive models like ARIMA \cite{5}, assuming linearity and stationarity, struggle with non-linear behaviors and rapid changes typical of our data. Prophet \cite{6} improves with seasonal adjustments but fails to handle irregular trends, while NeuralProphet \cite{7}, though better at managing non-linear trends, risks overfitting and is computationally intensive, with limited interpretability. Tree-based models such as XGBoost \cite{8} are effective but require extensive data and feature engineering, posing challenges with limited data. Feature-heavy methods like TS Fresh \cite{9} can introduce additional complexity and risk overfitting when data is scarce.
Recent advances in multidimensional time series analysis \cite{10} suggest that leveraging multiple dimensions can improve accuracy even with limited data. However, existing multidimensional approaches, such as ARIMAX \cite{11}, VAR-based models \cite{12}, and factor-based techniques \cite{13}, often rely on abundant data and emphasize latent factors or joint processes rather than considering the unique cohort-by-horizon structure. Similarly, cyclical and seasonal frameworks like SARIMA \cite{14} and TBATS \cite{15} capture periodic patterns along a single temporal axis but do not inherently incorporate a second dimension.

By bridging methodologies from traditional time series, multidimensional modeling, and cyclical frameworks, we aim to enhance forecasting accuracy across varying data sizes, ultimately narrowing the gap between the capacity of models to handle vast amounts of data and their effectiveness in constrained data scenarios.

\section{Proposed Method}
\subsection{Data Structure (Business Case)}

In business analytics, cohort-based data structures are prevalent for revenue analysis, characterized by two primary time dimensions. First, cohorts represent user groups segmented by their attribution time (usually users who installed the app in the same calendar month). The second dimension tracks the time since cohort attribution, with revenue measured for each cohort from the first month onward. This concept facilitates the collective analysis of user behavior and revenue over time, and can be represented as a matrix where rows correspond to different cohorts and columns represent successive time intervals since installation. The intersection of a row and a column within this matrix provides the cumulative revenue for a particular cohort at a specific time since installation.

A key characteristic is the diagonal division of known (historical) data above it and unknown (future) data below it, where each row is a cohort and each column represents time since installation. The proximity of data points to the diagonal indicates the prediction horizon, with those nearer representing a shorter forecast period.

This structure also incorporates a third time dependent element – the prediction time. With each progression in the calendar, the matrix is updated by introducing a new cohort to the matrix. The newest cohort typically presents the greatest challenge, offering limited historical data, yet requiring predictions over the longest horizon.

\subsection{Modeling}

Our database was structured in a tabular format, with each row representing a unique combination of cohort, time since installation, and revenue generated in the respective month. We also integrated features like subscription types, quantities, and revenues from first month post installation to provide additional context for each cohort.

The forecasting process involved repeatedly iterating over each column of the matrix, as detailed in Table \ref{tab:iteration}. For instance, during the initial iteration, we focused on the first column, which corresponds to one month post-event. Given that our forecast spans monthly intervals, this column typically included data from previous cohorts. We used these records along with their corresponding features to predict the single unknown record for that month. Subsequently, we incorporated the forecasted values from the former column into our dataset to enhance the feature set for predicting the outcomes in the next column. 

By adopting this approach, we maximized the use of known data at each stage, gradually building our predictions as we moved further. This method allowed for a dynamic and nuanced forecasting model capable of managing the complexities of our multidimensional data structure.

Traditional forecasting algorithms, often struggle due to the challenges posed by the scarcity of data for newer cohorts. Our approach is specifically designed to navigate this complexity. By leveraging the detailed insights provided by both the cohort and time-since-installation dimensions, our model effectively addresses the nuances of multidimensional data, making it well-suited for predicting future trends in user behavior and revenue generation.

\begin{table}[!ht]
\caption{Revenue Forecasting Modeling Iteration Process.*}
\label{tab:iteration}
\scriptsize
\begin{tabular}{clccccc}
\hline
\textbf{Iteration} & \textbf{Cohort} & \textbf{0} & \textbf{1} & \textbf{2} & \textbf{3} & \textbf{4} \\
\hline
1 & Sep 23 & 26,000\$ & 27,000\$ & 28,000\$ & 29,000\$ & 30,000\$ \\
1 & Oct 23 & 31,000\$ & 32,000\$ & 33,000\$ & 34,000\$ & unknown \\
1 & Nov 23 & 27,000\$ & 28,000\$ & 29,000\$ & unknown & unknown \\
1 & Dec 23 & 29,000\$ & 30,000\$ & unknown & unknown & unknown \\
1 & Jan 24 & 30,000\$ & \cellcolor[gray]{0.7}31,000\$ & unknown & unknown & unknown \\
\hline
2 & Sep 23 & 26,000\$ & 27,000\$ & 28,000\$ & 29,000\$ & 30,000\$ \\
2 & Oct 23 & 31,000\$ & 32,000\$ & 33,000\$ & 34,000\$ & unknown \\
2 & Nov 23 & 27,000\$ & 28,000\$ & 29,000\$ & unknown & unknown \\
2 & Dec 23 & 29,000\$ & 30,000\$ & \cellcolor[gray]{0.8}31,000\$ & unknown & unknown \\
2 & Jan 24 & 30,000\$ & \cellcolor[gray]{0.7}31,000\$ & \cellcolor[gray]{0.8}32,000\$ & unknown & unknown \\
\hline
3 & Sep 23 & 26,000\$ & 27,000\$ & 28,000\$ & 29,000\$ & 30,000\$ \\
3 & Oct 23 & 31,000\$ & 32,000\$ & 33,000\$ & 34,000\$ & unknown \\
3 & Nov 23 & 27,000\$ & 28,000\$ & 29,000\$ & \cellcolor[gray]{0.9 }30,000\$ & unknown \\
3 & Dec 23 & 29,000\$ & 30,000\$ & \cellcolor[gray]{0.8}31,000\$ & \cellcolor[gray]{0.9}32,000\$ & unknown \\
3 & Jan 24 & 30,000\$ & \cellcolor[gray]{0.7}31,000\$ & \cellcolor[gray]{0.8}32,000\$ & \cellcolor[gray]{0.9}33,000\$ & unknown \\
\hline  
\multicolumn{6}{l}{$^{*}$Note: This data is fictional and used solely for demonstration purposes.}
\end{tabular}
\end{table}

\subsection{Suggested Approach - 2D ARIMA}
In order to formulate the suggested 2D time series, we will use ARIMAX. The ARIMAX model contains an Auto-regressive component, Integrated component, Moving Average component and an Exogenous variables component. 

In the 2D time series approach, we will define \( \hat{Y}_{t,u} \) to be the predicted value at cohort  \( t \) and time since event  \( u \) for \( u = 1, 2, \ldots, U \). In our case, for 12 months prediction \( U = 12 \). 
In addition, we will define the complete series (column) of time since event \( u \) as the join of realized values and predicted values \( \tilde{Y}_{u} = [Y_{u}, \hat{Y}_{u}] \).

Then, the ARIMAX formulation for the 2D time series will be represented as:

\begin{align}
\hat{Y}_{t,u} = \mu_{u} + \sum_{i=1}^{p} \phi_{i,u} Y_{t-i,u} - \sum_{j=1}^{q} \theta_{j,u} \epsilon_{t-j,u} + \\ + \beta_{u-1} \tilde{Y}_{t,u-1} + \sum_{k=1}^{K} \beta_k X_{k,t} + \epsilon_{t,u} \nonumber
\end{align}

Where \( \tilde{Y}_{t,u-1} \) is the series at time since event  \( u - 1 \) (joint realized values and predicted), \( \beta_{u-1} \) is the coefficients of the previous time since event \( \tilde{Y}_{u} \),  \( X_{k,t} \) are the values of the exogenous variables at time \( t \), for each \( k \) exogenous variable represent cohort information and \( \beta_k \) are the coefficients of the exogenous variables.

In each iteration, we fit a model on the observed and the previous column’s observed and predictions, and by that we combine the cohort dimension (rows), the time from event dimension (column), and the time series dimension (diagonal).

In addition to this base process, there is an option to add more information as covariates (cohort information - e.g., number of users in the cohort). 

The advantage of this kind of approach is the lengthwise and crosswise combination of dimensions and rolling information from one iteration to another.

\subsection{Computational Efficiency}

While 2D time series modeling introduces additional computational overhead with time complexity of O(M · n · (p + q + d)) compared to traditional O(n · (p + q + d)) approaches, it offers improved performance by capturing complex seasonal and cohort effects, with the computational trade-off justified by enhanced forecasting capabilities.

\section{Evaluation}

 To underscore the effectiveness of the 2D time series model, we conducted a comprehensive comparative analysis against well-established models, including Linear Regression, XGBoost, and Prophet. We excluded DL methods due to their substantial data requirements, which our datasets do not meet \cite{16}.
 
\subsection{Data and Modeling Methodology}

We evaluate performance and generalizability using two distinct datasets: 
\begin{enumerate}
  \item \textit{Applications Historical Revenue Data:} App subscriptions from January 2020 to February 2023, covering unique cohort-app-geo-location combinations.
  \item \textit{Customer Cohorted Purchase Records:} Monthly and yearly subscriptions segmented by gender from January 2017 to December 2021 \cite{17}. 
\end{enumerate}

A key feature of our methodology was the simulation of monthly prediction, structured as follows:
\begin{itemize}
\item \textit{Forecast Horizon}: At the beginning of each prediction month during the forecast period, the models forecast revenue for future months of all corresponding cohorts.
\item \textit{Cohorts}: Each prediction extended to cover up to one year after installation.
\item \textit{Recurring Evaluations}: This process was replicated at the start of each month throughout the study. This allowed for an ongoing assessment of the models under varying conditions such as changing marketing strategies, emerging market trends, and global events.
\end{itemize}

\subsection{Evaluation Methodology}
To thoroughly evaluate and compare models performance, we selected several evaluation metrics, each chosen to highlight different aspects of forecasting accuracy and robustness. Using multiple metrics ensures a comprehensive assessment that captures the nuances of the models behavior.

The evaluation metrics were: 
\begin{itemize}
    \item Root Mean Square Error (RMSE)
    \item Mean Absolute Error (MAE)
    \item Symmetric Mean Absolute Percentage Error (sMAPE)
\end{itemize}

These metrics together provide a well-rounded evaluation of model accuracy and error distribution, reflecting both the magnitude and consistency of errors across predictions, and offering a multifaceted assessment framework for assessing forecasting performance \cite{18,19,20,21,22,23}.

\begin{table}[!ht]
\caption{Forecasting Performance Comparison for Both Datasets}
\label{tab:combined_performance_comparison}
\centering
\begin{adjustbox}{max width=\linewidth}
\begin{tabular}{llcccc}
\toprule
\textbf{Dataset} & \textbf{Metric} & \textbf{2D Model} & \makecell{\textbf{Linear}\\\textbf{Regression}} & \textbf{XGBoost} & \textbf{Prophet}\\
\midrule
 & MAE   & \textbf{0.06 $\pm$ 0.08} & 0.28 $\pm$ 0.27 & 1.10 $\pm$ 0.45 & 1.07 $\pm$ 2.45 \\
Applications & RMSE  & \textbf{0.24 $\pm$ 0.28} & 0.53 $\pm$ 0.52 & 1.05 $\pm$ 0.67 & 1.03 $\pm$ 1.57 \\
 & sMAPE & \textbf{6.45 $\pm$ 8.33} & 27.32 $\pm$ 22.62 & 182.66 $\pm$ 75.96 & 51.85 $\pm$ 84.99 \\
\midrule
 & MAE   & \textbf{0.03 $\pm$ 0.04} & 0.19 $\pm$ 0.26 & 0.07 $\pm$ 0.10 & 1.07 $\pm$ 2.53 \\
Costumer Subscription & RMSE  & \textbf{0.17 $\pm$ 0.20} & 0.44 $\pm$ 0.51 & 0.27 $\pm$ 0.19 & 1.03 $\pm$ 1.59 \\
 & sMAPE & \textbf{3.28 $\pm$ 4.15} & 24.39 $\pm$ 26.15 & 7.87 $\pm$ 10.04 & 52.81 $\pm$ 94.43 \\
\bottomrule
\end{tabular}
\end{adjustbox}
\end{table}

The efficacy of the 2D model, demonstrated across both datasets and a range of metrics, is summarized in Table \ref{tab:combined_performance_comparison}. These results highlight the model's superior forecasting capabilities, consistently outperforming traditional models in both accuracy and reliability, making it a valuable tool for advanced forecasting across various industry sectors.

Following the multi-metric evaluation, Linear Regression (LR) emerged as the best-performing model and was selected as the baseline for further detailed comparisons. In these analyses, sMAPE  employed as the evaluation metric due to its robustness in handling outliers and its ability to provide a normalized measure of error.

In our comparative analysis, we place particular focus on the longest horizon prediction of 1-year ahead for the most recent cohort. These predictions are especially challenging due to limited historical data and the longer forecasting horizon. By that we rigorously evaluate the model’s performance under the most demanding conditions. 

\begin{figure}
    \vspace{-3pt}
    \centering
    \includegraphics[width=1\linewidth]{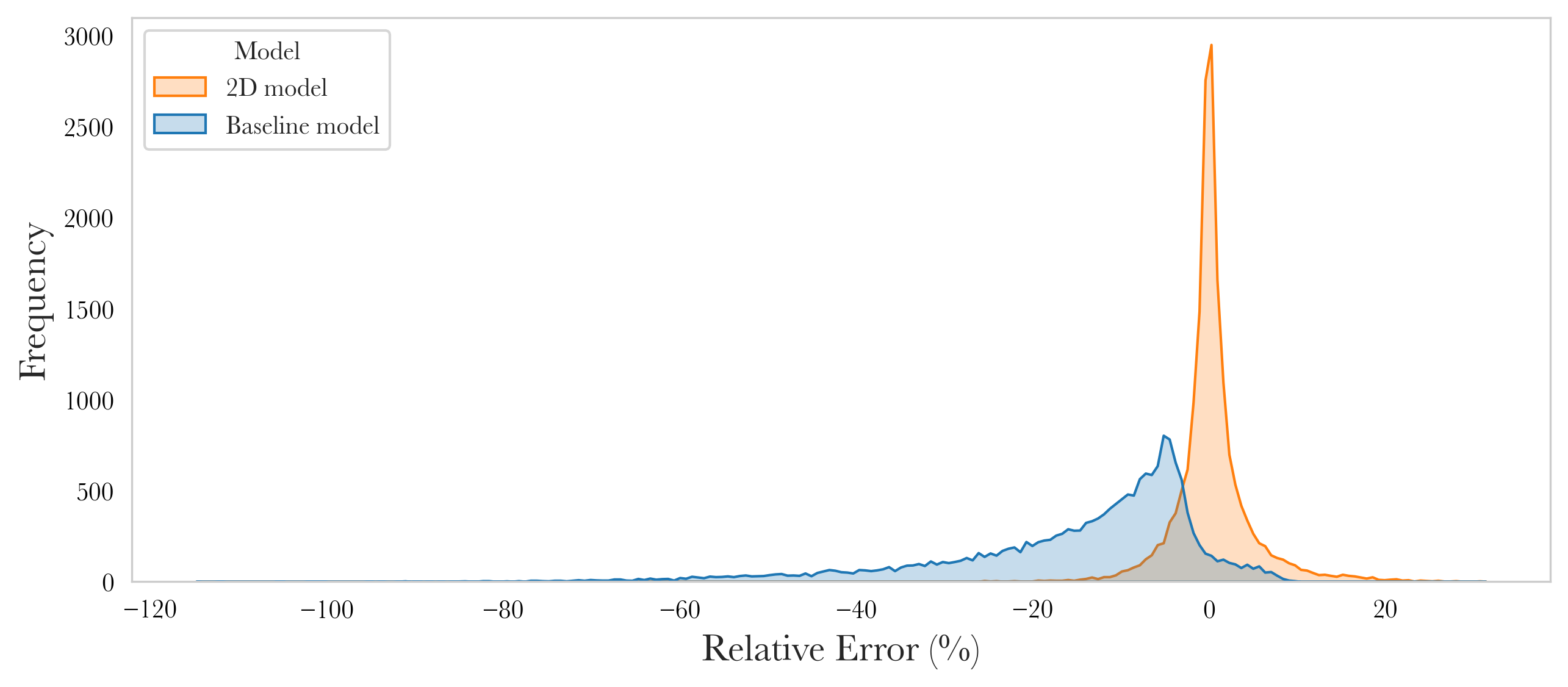}
    \vspace{-3pt}
    \caption{Relative errors Distribution. Most errors cluster around 0\%, indicating accurate predictions, with a few deviations, especially in the negative direction for Baseline model.}
    \label{fig:hist}
\end{figure}

\begin{figure}
    \vspace{-3pt}
    \centering
    \includegraphics[width=1\linewidth]{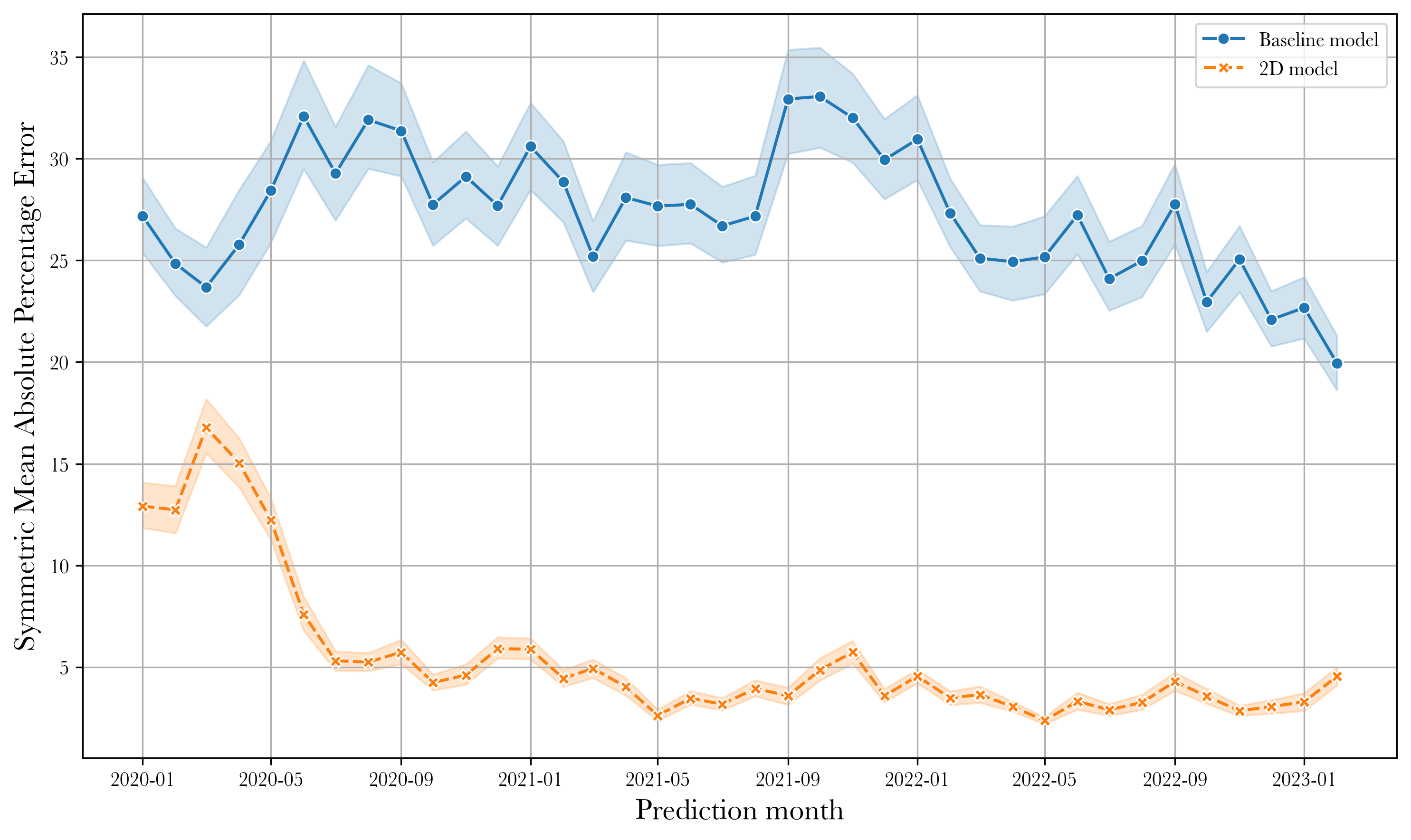}
    \vspace{-3pt}
    \caption{Median Absolute Percentage Error (sMAPE) over time. The Baseline Model displays higher errors and errors variability, while the 2D Model remains relatively consistent.}
    \label{fig:sMAPE error}
\end{figure}

Our statistically significant findings consistently demonstrate the 2D model's superior accuracy across all applications and market segments. The 1-year horizon results further emphasize the 2D model’s strength, with significantly lower sMAPE scores compared to the Baseline model. This is especially significant given the challenges associated with long-term forecasting and the limited data available for such predictions.

\begin{figure}[ht]
    \vspace{-3pt}
    \centering
    \includegraphics[width=1\linewidth]{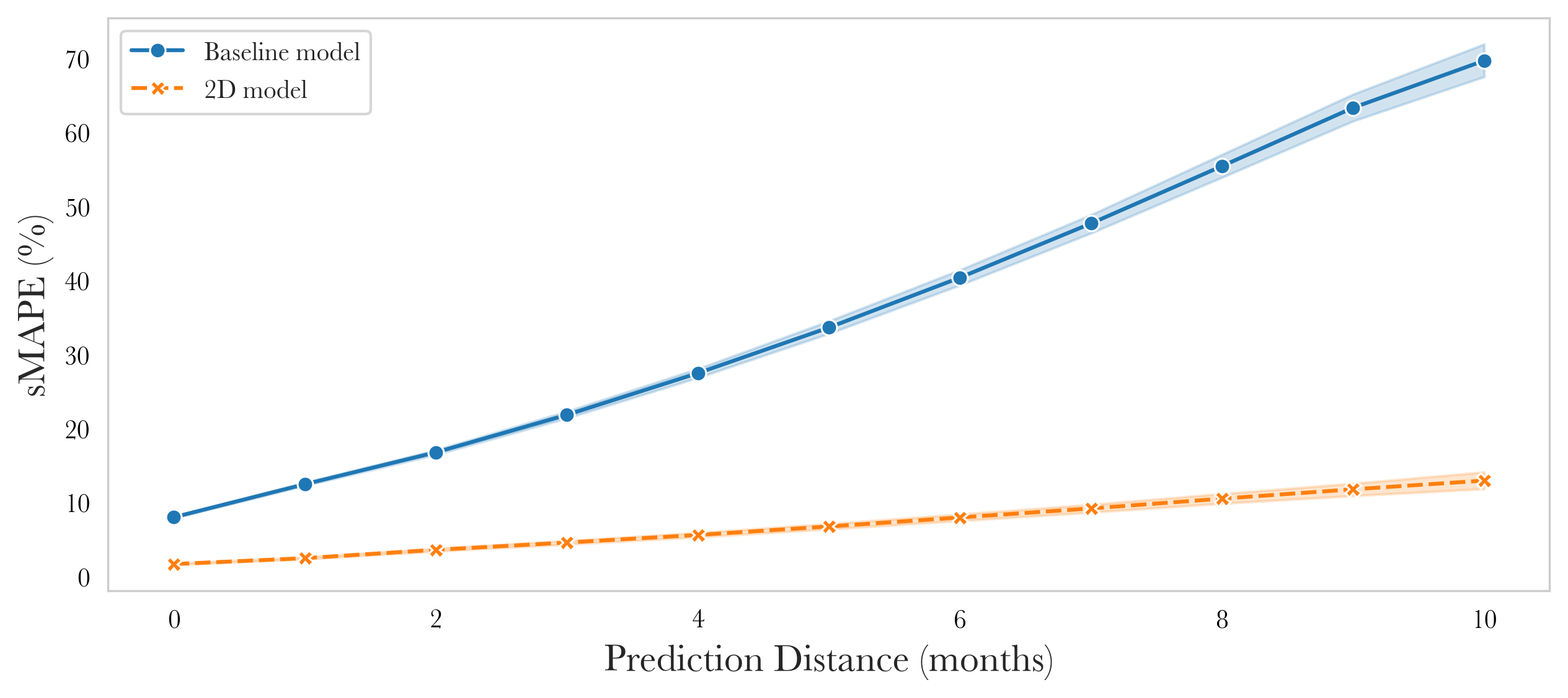}
    \vspace{-3pt}
    \caption{Symmetric Mean Absolute Percentage Error (sMAPE) over varying prediction distances. The 2D model exhibits a consistent error rate, while the Baseline model shows increasing error with extended prediction steps.}
    \label{fig:error_over_distance}
\end{figure}

The 2D model exhibited several advantageous characteristics, including the ability to quickly adapt to changes and minimize errors over time. As shown in Figures \ref{fig:sMAPE error} and \ref{fig:error_over_distance}, the model's adaptability was particularly evident in its responsiveness to market fluctuations, maintaining low forecast error. The 2D model sustained a relatively consistent error rate, indicating a stable prediction range. In contrast, the Baseline model showed a trend of increasing error with longer prediction distances. Additionally, the model's flexibility in adjusting forecast horizon lengths and its minimal requirement for historical data make it highly suited for dynamic, fast-changing markets. These combined attributes—adaptability, accuracy, and flexibility, set the 2D model apart from traditional forecasting methods.

\subsection{Results summary}
The 2D model exhibits exceptional adaptability to user behavior changes, maintaining relevance and accuracy in rapidly shifting conditions. Its capacity for stable, reproducible predictions over time, along with its flexibility in adjusting forecast horizons, further underscores its robustness. Notably, the 2D model demonstrates a consistent trend in reducing high errors over subsequent months, showcasing its efficacy in refining predictions.

\section{Conclusions}

\subsection{Overview}

In this study, we propose an innovative 2D time-series modeling approach for time series forecasting, specifically designed to address challenges in scenarios with limited historical data and multidimensional data structures, such as those common in user subscription-based businesses. Our analysis demonstrates the utility of the 2D model in enhancing the precision of future revenue forecasts in this context, compared to small time series SOTA models.

\subsection{Key Findings}

Our findings highlight several key advantages of the 2D model:
\begin{itemize}[leftmargin=*]
    \item \textit{Enhanced forecasting accuracy}: The 2D time-series model consistently outperforms the Baseline model, demonstrating lower mean sMAPE and higher precision across various applications, geographic locations, and market segments.
    \item \textit{Long-term forecasting capabilities}: The 2D model excels in long-term forecasting, particularly noticeable in 1-year horizon sMAPE results, highlighting its suitability for strategic business planning.
    \item \textit{Stability and flexibility}: With stable error reduction over time and adaptability to different forecasting horizons, the model's robustness and precision make it highly effective in dynamic market conditions.
\end{itemize}

\subsection{Theoretical and Practical Implications}

This work expands the field of multidimensional time series forecasting by addressing the complexities found in small datasets with varying dimensions and offering a new perspective in time series analysis.

Practically, our 2D time-series model serves as a powerful tool for businesses, particularly in subscription-based services. While its immediate application is in the mobile market, the model’s utility extends across sectors, making it valuable for strategic planning and operations, especially in emerging or fast-changing markets where data might be scarce or becomes outdated quickly.

\subsection{Real-World Implementation} Our model has been implemented in production for real-time tasks and successfully tested for up to 32 future steps. This range may not represent the model’s full potential, suggesting it could perform well beyond this under different conditions. Its consistent performance in real-world scenarios confirms the model's robustness and adaptability, demonstrating superiority over both traditional and advanced methods in complex forecasting situations. These practical applications validate our experimental results and highlight the model’s potential as a reliable tool in various industries.

\subsection{Limitations and Future Research}

While demonstrating promising potential, It should be taken into consideration that the iterative approach, limits computational parallelization and introduces complexities in uncertainty estimation.
Future research should focus on integrating external factors and validating the model across diverse industries to comprehensively assess its broader applicability and refine its predictive capabilities.

\bibliographystyle{IEEEtran}

\begin{thebibliography}{99}

\bibitem{1}
F. Dama and C. Sinoquet, "Analysis and modeling to forecast in time series: A systematic review" \emph{arXiv preprint arXiv:2104.00164}, 2021.

\bibitem{2}
R. Adhikari and R. K. Agrawal, "An introductory study on time series modeling and forecasting" \emph{arXiv preprint arXiv:1302.6613}, 2013.

\bibitem{3}
B. Lim and S. Zohren, "Time-series forecasting with deep learning: A survey" \emph{Philosophical Transactions of the Royal Society A}, vol. 379, no. 2194, pp. 20200209, 2021.

\bibitem{4}
C. Faloutsos, J. Gasthaus, T. Januschowski, and Y. Wang, "Forecasting big time series: Old and new" \emph{Proceedings of the VLDB Endowment}, vol. 11, no. 12, pp. 2102--2105, 2018.

\bibitem{5}
T. Hastie, "The elements of statistical learning: Data mining, inference, and prediction," Springer, 2009.

\bibitem{6}
S. J. Taylor and B. Letham, "Forecasting at scale" \emph{The American Statistician}, vol. 72, no. 1, pp. 37--45, 2018.

\bibitem{7}
O. Triebe et al., "NeuralProphet: Explainable forecasting at scale" \emph{arXiv preprint arXiv:2111.15397}, 2021.

\bibitem{8}
T. Chen and C. Guestrin, "XGBoost: A scalable tree boosting system" in \emph{Proc. 22nd ACM SIGKDD Int. Conf. Knowledge Discovery and Data Mining}, 2016, pp. 785--794.

\bibitem{9}
M. Christ, N. Braun, J. Neuffer, and A. W. Kempa-Liehr, "Time series feature extraction on basis of scalable hypothesis tests (tsfresh--A Python package)" \emph{Neurocomputing}, vol. 307, pp. 72--77, 2018.

\bibitem{10}
X. Chen and L. Sun, "Low-rank autoregressive tensor completion for multivariate time series forecasting" \emph{arXiv preprint arXiv:2006.10436}, 2020.

\bibitem{11}
N. S. Arunraj, D. Ahrens, and M. Fernandes, "Application of SARIMAX model to forecast daily sales in food retail industry" \emph{International Journal of Operations Research and Information Systems}, vol. 7, no. 2, pp. 1--21, 2016.

\bibitem{12}
H. Lütkepohl, \emph{New Introduction to Multiple Time Series Analysis}. Berlin, Germany: Springer Science \& Business Media, 2005.

\bibitem{13}
J. Bai and S. Ng, "Determining the number of factors in approximate factor models" \emph{Econometrica}, vol. 70, no. 1, pp. 191--221, 2002.

\bibitem{14}
G. E. P. Box, G. M. Jenkins, G. C. Reinsel, and G. M. Ljung, \emph{Time Series Analysis: Forecasting and Control}. John Wiley \& Sons, 2015.

\bibitem{15}
A. M. De Livera, R. J. Hyndman, and R. D. Snyder, "Forecasting time series with complex seasonal patterns using exponential smoothing" \emph{Journal of the American Statistical Association}, vol. 106, no. 496, pp. 1513--1527, 2011.

\bibitem{16}
C. Zhang, N. N. A. Sjarif, and R. Ibrahim, "Deep learning models for price forecasting of financial time series: A review of recent advancements: 2020--2022" \emph{Wiley Interdisciplinary Reviews: Data Mining and Knowledge Discovery}, vol. 14, no. 1, pp. e1519, 2024.

\bibitem{17}
"Customer subscription data" Kaggle, 2022. [Online]. Available: https://www.kaggle.com/datasets/gsagar12/dspp1

\bibitem{18}
J. S. Armstrong and F. Collopy, "Error measures for generalizing about forecasting methods: Empirical comparisons" \emph{International Journal of Forecasting}, vol. 8, no. 1, pp. 69--80, 1992.

\bibitem{19}
P. Goodwin and R. Lawton, "On the asymmetry of the symmetric MAPE" \emph{International Journal of Forecasting}, vol. 15, no. 4, pp. 405--408, 1999.

\bibitem{20}
L. Ren and Y. Glasure, "Applicability of the revised mean absolute percentage errors (MAPE) approach to some popular normal and non-normal independent time series" \emph{International Advances in Economic Research}, vol. 15, pp. 409--420, 2009.

\bibitem{21}
A. De Myttenaere, B. Golden, B. Le Grand, and F. Rossi, "Mean absolute percentage error for regression models" \emph{Neurocomputing}, vol. 192, pp. 38--48, 2016.

\bibitem{22}
C. Tofallis, "A better measure of relative prediction accuracy for model selection and model estimation" \emph{Journal of the Operational Research Society}, vol. 66, no. 8, pp. 1352--1362, 2015.

\bibitem{23}
B. E. Flores, "A pragmatic view of accuracy measurement in forecasting" \emph{Omega}, vol. 14, no. 2, pp. 93--98, 1986.


\end{thebibliography}

\setlength{\abovedisplayskip}{2pt}
\setlength{\belowdisplayskip}{2pt}
\setlength{\abovedisplayshortskip}{2pt}
\setlength{\belowdisplayshortskip}{2pt}

\end{document}